\pdfoutput=1 
\documentclass[letterpaper]{article} 
\usepackage{aaai23}  
\usepackage{times}  
\usepackage{helvet}  
\usepackage{courier}  
\usepackage[hyphens]{url}  
\usepackage{graphicx} 
\usepackage{natbib}  
\usepackage{caption} 
\usepackage{algorithm}
\usepackage{algorithmic}

\usepackage{newfloat}
\usepackage{listings}
\DeclareCaptionStyle{ruled}{labelfont=normalfont,labelsep=colon,strut=off} 
\floatstyle{ruled}
\newfloat{listing}{tb}{lst}{}
\floatname{listing}{Listing}
\usepackage{fontenc}

\usepackage{xspace}
\usepackage{xcolor}
\usepackage{booktabs}
\usepackage[inline]{enumitem}
\usepackage{amsmath}
\usepackage{comment}
\usepackage{multirow}
\usepackage{subcaption}
\usepackage{cleveref}
\usepackage{siunitx}

\author{
    Chieh-Yang Huang,\textsuperscript{\rm 1}~
    Saniya Naphade,\textsuperscript{\rm 2}{\equalcontrib}~
    Kavya Laalasa Karanam,\textsuperscript{\rm 3}{\equalcontrib}\\
    Ting-Hao `Kenneth' Huang\textsuperscript{\rm 1}
}
\affiliations{
    \textsuperscript{\rm 1} Pennsylvania State University, University Park, PA, USA. \{chiehyang,txh710\}@psu.edu \\
    \textsuperscript{\rm 2} GumGum Inc., Los Angeles, CA, USA. saniya.naphade@gmail.com \\
    \textsuperscript{\rm 3} Intel Corporation, Santa Clara, CA, USA. laalasa.kar@gmail.com \\

}

\title{Conveying the Predicted Future to Users: A Case Study of Story Plot Prediction}

\newcommand{\cy}[1]{{\small\textcolor{orange}{\bf [#1 --CY]}}}

\newcommand{\eg}{{\it e.g.}}
\newcommand{\ie}{{\it i.e.}}

\newcommand{\smallpvalue}[0]{\textless0.001}

\begin{document}

\maketitle

\pagestyle{plain}

\begin{abstract}
Creative writing is hard: Novelists struggle with writer's block daily. 
While automatic story generation has advanced recently, it is treated as a ``toy task'' for advancing artificial intelligence rather than helping people.
In this paper, we create a system that produces a short description that narrates a predicted plot using existing story generation approaches.
Our goal is to assist writers in crafting a consistent and compelling story arc.
We conducted experiments on Amazon Mechanical Turk (AMT) to examine the quality of the generated story plots in terms of consistency and \textit{storiability}.
The results show that short descriptions produced by our frame-enhanced GPT-2 (FGPT-2) were rated as the most consistent and \textit{storiable} among all models; FGPT-2's outputs even beat some random story snippets written by humans.
Next, we conducted a preliminary user study using a story continuation task where AMT workers were given access to machine-generated story plots and asked to write a follow-up story.
FGPT-2 could positively affect the writing process, though people favor other baselines more.
Our study shed some light on the possibilities of future creative writing support systems beyond the scope of completing sentences.
Our code is available at: \url{https://github.com/appleternity/Story-Plot-Generation}.

\end{abstract}

\section{Introduction}


\begin{figure}[t]
    \centering    
    \includegraphics[width=0.8\columnwidth]{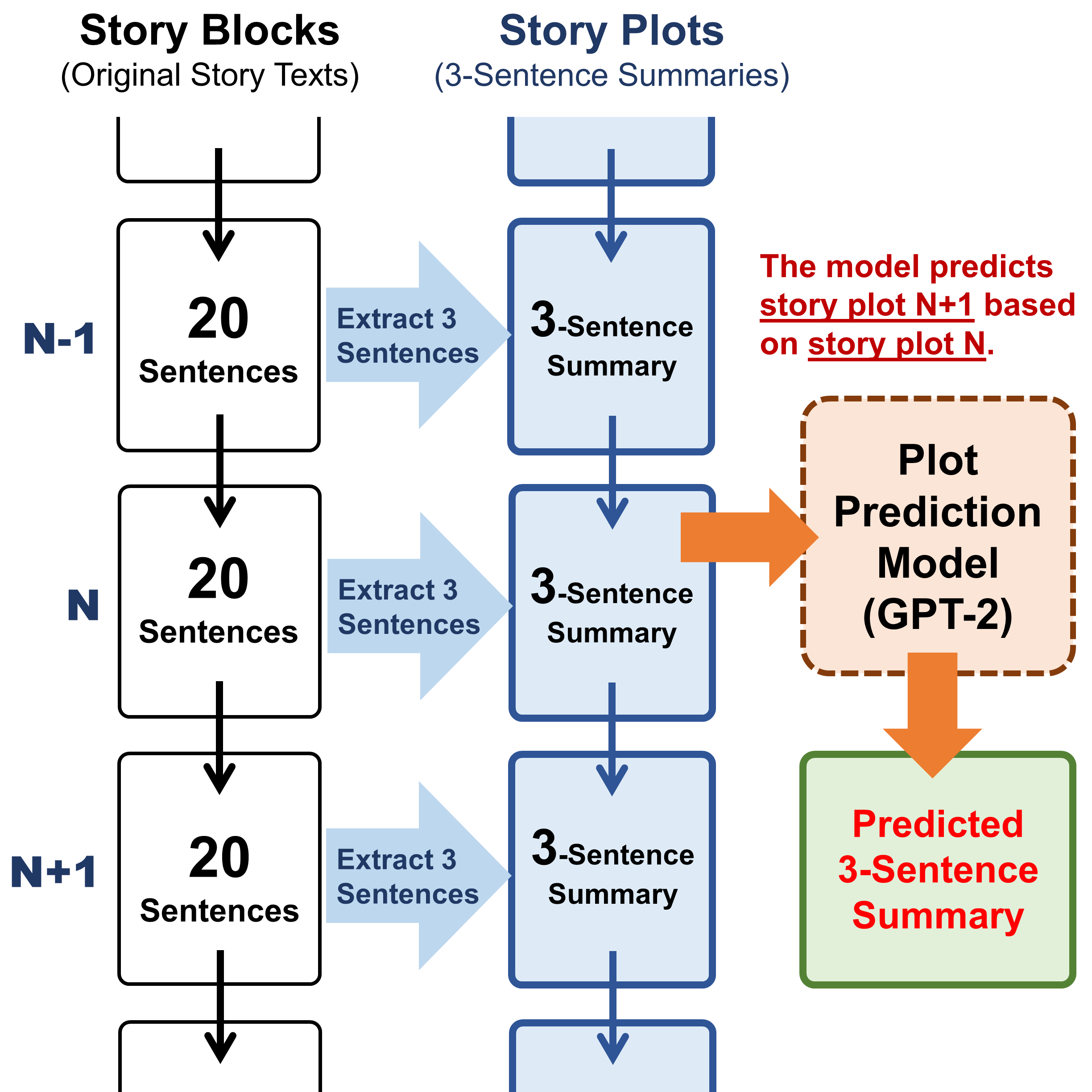}
    \caption{We view a long novel as a sequence of fixed-sized story blocks. The goal of the proposed task is to consider the previous story block (\ie, $B_{n}$) and generate a short description for the next story block (\ie, $B_{n+1}$). We define a short description as a three-sentence summary of a story block.}
    \label{fig:overview}
 \end{figure}


Storytelling is an important human activity.
People engage in storytelling to communicate, teach, entertain, establish identity, or relate to each other in meaningful ways.
However, creative writing is known to be a cognitively demanding task, and writers struggle with writer's block daily.
Researchers and the industry have created a series of techniques that support human writing.
Many techniques focus on lower-level language support, such as auto-completion, grammar checking, or typo detection, and these have proven helpful and are widely used.
On the other hand, the techniques aiming to provide higher-level guidance, such as story generation, have long been treated only as in-the-lab artificial intelligence tasks.
Automatic story generation, for example, was primarily developed and tested using toy datasets composed of stories that are 
{\em (i)} extremely short (for example, containing five sentences, such as ROCStories~\cite{mostafazadeh-etal-2016-corpus}), 
{\em (ii)} written under artificial constraints to make it easier for machines to learn (\eg, GLUCOSE~\cite{mostafazadeh-etal-2020-glucose}), or 
{\em (iii)} based on the assumption of a story starter prompt (\eg, WritingPrompt~\cite{fan-etal-2018-hierarchical}).
However, real-world writers compose novels with over 10,000 words, work with blank pages with few constraints, and can get stuck anywhere in the middle of a draft.
As the models trained on toy datasets inevitably generate stories inheriting the data's characteristics, it is unclear how well modern story generation models can be used to support writers in practice.

In this paper, we aim to support creative writing in practical terms. 
We view a long novel as a sequence of fixed-sized story blocks (\eg, 20 sentences).
The goal is to generate a short description that narrates future story plots for the next story block (\ie, $B_{n+1}$) using the previous story block (\ie, $B_{n}$).
We define a story plot as a short summary over a story block that illustrates the key follow-up idea instead of the detailed full text.
Three existing story generation models, 
Fusion-based seq2seq~\cite{fan-etal-2018-hierarchical}, 
Plan-and-Write~\cite{yao2019plan}, and 
GPT-2~\cite{radford2019language} enhanced with semantic frame representation~\cite{huang-huang-2021-semantic}, 
are adapted to predict the follow-up story plot given the context of the previous story block.

We first conduct a quality assessment study on Amazon Mechanical Turk (AMT) to measure the quality of the machine-generated story plots.
In this study, crowd workers are recruited to 
{\em (i)} read a previous story block and six follow-up story plots, and 
{\em (ii)} rank the quality in terms of consistency and storiability~\cite{roemmele2021inspiration}.
The experiment shows that story plots generated by our frame-enhanced GPT-2 are more consistent than randomly selected plots written by humans and are competitive with them in the sense of storiability.
The result suggests that human-written plots are still strong baselines, especially the ground truth, but frame-enhanced GPT-2 is capable of generating consistent and storiable story plots to a certain level.

We further conduct a writing task study on AMT to understand how much humans can benefit from machine-generated plots.
In this study, crowd workers are asked to develop a 100-word follow-up story given the previous story block and the four follow-up story plots as hints.
After finishing the writing task, we collect crowd workers' self-reported judgments on four aspects: degree of inspiration, helpfulness, readability, and creativity. 
The result shows that frame-enhanced GPT-2 produces output that is less inspiring when compared to strong baselines such as ground truth and GPT-3. 
However, analyses of the written stories also suggest that, despite being less favored by humans, frame-enhanced GPT-2 still has a positive influence on the written story draft.
This finding also echoes \citet{roemmele2021inspiration}'s inspiration-through-observation paradigm:
Human writing can still be improved even with less storiable machine-generated texts.

\section{Related Work}

Our work is mainly related to {\em (i)} supporting creative writing and {\em (ii)} story generation.

\subsection{Supporting Creative Writing}

Prior research has supported creative writing in different ways.
InkWell mimics a specified writer's personality traits and revises the draft to provide stylistic variations~\cite{gabriel2015inkwell}.
Metaphoria generates metaphorical connections according to the user's input to help create metaphors~\cite{10.1145/3290605.3300526}.
Creative Help generates a follow-up sentence as a suggestion for creative writing using a recurrent neural network~\cite{roemmele2015creative}.
Heteroglossia collects story plot ideas using a crowdsourcing approach to support writers in continuing the story when stuck due to writer's block~\cite{huang2020heteroglossia}.
Scheherazade is built for interactive narrative generation with a crowd-powered system to collect narrative examples~\cite{li2015scheherazade}. 
\citet{Clark:2018:CWM:3172944.3172983} explore the process of machine-in-the-loop creative writing and find that machine-generated suggestions should achieve a balance between coherency and surprise.
\citet{roemmele2021inspiration} studies the inspiration-through-observation paradigm and finds that people produce appealing sentences when observing the generated examples.
Compass identifies and fills in unnoticed missing information in stories~\cite{DBLP:journals/corr/abs-2202-13151}.
\citet{padmakumar-he-2022-machine} build a machine-in-the-loop system to rewrite a span of text or fill in sentences between two pieces of text when requested.



Recently, large language models (LLMs) have shown incredible power in text continuation, rewriting, few-shot learning, and so on. 
Many researchers have explored how LLMs can be used to support creativity. 
Storium fine-tunes GPT-2 to consider complicated contextual information (intro, character, and so on) to generate a few follow-up sentences to continue the story~\cite{akoury-etal-2020-storium}.
Story Centaur provides an interface where users can provide few-shot learning examples to teach LLMs new functions~\cite{swanson-etal-2021-story}. 
CoPoet, a collaborative poetry writing system, allows users to control an LLM by specifying the attributes of the desired text~\cite{chakrabarty2022help}. 
TaleBrush allows users to control a protagonist's fortune  through a line sketching interaction, and the user-specified fortune curve is used to guide an LLM's story generation process~\cite{chung2022talebrush}.
CoAuthor supports writing by providing a sentence to continue the given draft by GPT-3~\cite{10.1145/3491102.3502030}. 
Sparks inspires scientific writing by using LLMs to generate sentences that could spark users' ideas~\cite{10.1145/3532106.3533533}. 
Wordcraft allows users to interact with LLMs through a chatbot interface~\cite{ippolito2022creative}.
Dramatron, built with an LLM with prompt-chaining mechanism, could write theatre scripts and screenplays together with users~\cite{mirowski2022co}.
Unlike most of the prior works, where generated sentences are ready to use in the story, our work aims to generate a short summary for the follow-up story and expects users to develop exact story content manually.








\subsection{Story Generation}

Traditional story generation focuses on producing logically coherent stories using planning or reasoning-based approaches~\cite{riedl2010narrative,li2013story}.
Recently, neural story generation models~\cite{peng-etal-2018-towards,fan-etal-2018-hierarchical} and pre-trained models~\cite{radford2019language,keskar2019ctrl} have been used for story generation in an end-to-end manner.
However, these models still suffer from the issue of generating repetitive and insufficiently diverse stories~\cite{see-etal-2019-massively}.
To further enhance the coherence among sentences and events, researchers design a variety of intermediate representations to guide the story generation process, including 
event triplets~\cite{martin2018event}, 
keyword storylines~\cite{yao2019plan}, 
critical phrases~\cite{xu2018skeleton}, 
action plans with semantic role labeling (SRL)~\cite{fan-etal-2019-strategies}, 
content planning (keyphrase and sentence-level position)~\cite{hua2020pair}, and 
plot structure based on SRL~\cite{goldfarb-tarrant-etal-2020-content}.
However, most of these work on short stories, such as WritingPrompt~\cite{fan-etal-2018-hierarchical}, ROCStories~\cite{mostafazadeh-etal-2016-corpus}, or WikiPlots~\cite{bamman-etal-2013-learning}.
Unlike real-world novels-- which usually have more than 10,000 words-- the stories from these datasets often end up with under 1,000 or even 100 words.

\section{Plot Prediction}


We follow \citeauthor{huang-huang-2021-semantic} to split a full story into a sequence of story blocks 
and each story block contains a fixed number of sentences.
Note that the size of the story block can vary to fulfill different purposes.
Large story blocks (200 sentences or beyond) can capture the high-level ideas among chapters; whereas small story blocks (five or ten sentences) can be used to model the event relationships in a near future.
In this paper, we focus on medium size story blocks (20 sentences).

Next, we define \textbf{a story plot} as \textbf{a three-sentence summary} of a huge story block. 
The plot prediction task, thus, is defined as using the story plot in story block \textit{n}
to predict the story plot in story block \textit{n+1}.
In this section, we first describe how we collect the story plots using the extractive summarization model;
and then detail how we adapt the three existing story generation models to our problem.

\subsection{Collecting Story Plots}
To generate such a summary for every story block, we use Matchsum~\cite{zhong2020extractive}, an extractive summarization model. 
We train the Matchsum model on the Booksum dataset~\cite{kryscinski2021booksum} where each paragraph is paired with a one- or two-sentence summary.
To ensure the training instances from Booksum are similar to those in our story block setup (20 sentences), 
only 20,709 paragraphs with more than 10 sentences are kept for training.
The fine-tuned Matchsum model is then applied to our Bookcorpus dataset to generate the story plot for all the 900k story blocks.
We randomly select 1k instances as the validation set in order to observe if the model converges or not in the training process.


\subsection{Story Plot Generation Models}
\label{sec:story-plot-generation-model}
Here, we adapt three existing models to our story plot generation task:
(\textit{i}) Fusion-based Seq2Seq~\cite{fan-etal-2018-hierarchical},
(\textit{ii}) Plan-and-Write~\cite{yao2019plan},
and (\textit{iii}) GPT-2~\cite{radford2019language} guided by semantic frame representation~\cite{huang-huang-2021-semantic}.

\subsubsection{Fusion-Based Seq2seq.}
\label{sec:model-fusion-seq2seq}
The fusion-based mechanism~\cite{fan-etal-2018-hierarchical} is a hierarchical model 
where a seq2seq model is trained on top of a pre-trained seq2seq model.
The underlying Convolutional Seq2Seq model is trained on a premise or prompts, 
the plot of story block \textit{n} in our case.
The fusion model, another convolutional seq2seq model is then trained 
on top of the previous seq2seq model to encourage the model 
to focus on the link between the prompt and the generated story, 
making it easier to generate consistent stories and reducing the tendency to drift off-topic. 
Given the prompt, the model generates one of the possible directions 
story block \textit{n+1} in which the story could progress further.

We tokenize the plots using NLTK, 
turn words that appear less than 10 times in the training set to \texttt{<UNK>},
and follow the paper's default hyper-parameters to train the model~\cite{fan-etal-2018-hierarchical}.
The base seq2seq model is trained for 20 epochs and then used to train the fusion model which takes 15 epochs.
We use top-k sampling to generate story plots with k = 100, temperature = 0.8, and unknown token penalty = 100.
Plot lengths are limited to 31 to 71 tokens as the average length in the training set is 51.08.


\subsubsection{Plan-and-Write (P\&W).}
\label{sec:plan-and-write}
Plan-and-Write~\cite{yao2019plan} makes use of static planning which 
generates storylines as a standard intermediate representation to create coherent and diverse stories. 
Storylines are depicted as a sequence of important words that estimate structures for a real story plot. 
The P\&W model takes a prompt as the input to 
(\textit{i}) first plan the storylines and (\textit{ii}) then generate the whole story.
Following \citet{yao2019plan}'s setup, we apply RAKE algorithm~\cite{RAKE2010} to extract keywords 
from the plot of story block \textit{n+1} to form the storylines.

We tokenize the plots using NLTK and 
turn words that appear less than 10 times in the training set to \texttt{<UNK>}.
The storyline generation model is based on a 3-layer LSTM with embedding size = 300, hidden size = 300;
and the plot generation model is based on a 5-layer LSTM with embedding size = 512, hidden size = 512;
We use Adam~\cite{kingma2014adam} as the optimizer to train the model.
For the rest of the hyper-parameters, we follow the setting in the original paper.
The storyline model is trained for 100 epochs and the plot generation model is trained for 40 epochs.
We use temperature sampling to generate storylines and the final story plots with temperature = 0.8.
Plan-and-Write does not handle unknown tokens so we add our implementation by setting the sampling probability of \texttt{<UNK>} to zero.
Again, plot lengths are limited to 31 to 71 tokens.
After obtaining the generated story plots, to remove the artifact,
we further apply the Treebank detokenizer~\cite{BirdKleinLoper09} to remove extra spaces and Truecase~\cite{lita2003truecasing} to capitalize necessary letters.


\subsubsection{Frame-enhanced GPT-2 (FGPT-2).}
\label{sec:gpt-2-model}
Semantic frame representations~\cite{huang-huang-2021-semantic} encode high-level semantic units into vectors using their corresponding importance, which has been shown to be effective for representing a longer context.
We take advantage of the semantic frame representation to encode longer contextual information and \citet{huang-huang-2021-semantic}'s semantic frame forecast model to generate the guidance of the follow-up story block. 
Note that the predicted frame representation contains semantic units that are expected to happen in the next story block which serves as a \textbf{goal-setting} function in the writing process theory~\cite{flower1981cognitive}.
We build a sequence-to-sequence model using the pre-trained GPT-2 model where two GPT-2 models are used for the encoder and decoder respectively. 
The frame representation of story block \textit{n} and the predicted frame representation of story block \textit{n+1} is passed through a linear layer to fit the GPT-2's word embedding dimension and inputted into GPT-2 encoder as two words.
The encoder input can be described as:
$
    \mathbf{X} = [ \mathbf{x}_1, \mathbf{x}_2, \cdots, \mathbf{x}_n, \mathbf{f}_{n}, \hat{\mathbf{f}}_{n+1} ]
$
where $\mathbf{x}_i$ is the $i$-th word in the story plot of story block \textit{n},
$\mathbf{f}_{n}$ is the adapted frame representation of story block \textit{n},
and $\hat{\mathbf{f}}_{n+1}$ is the LGBM's prediction of the frame representation for story block \textit{n+1}.
We added the \texttt{START} and \texttt{<PAD>} tokens to the model to enable the sequence-to-sequence training framework.
The sequence $[$\texttt{<START>}$, y_1, y_2, \cdots, y_n,$\texttt{<|endoftext|>}$]$, 
where $y_i$ is the $i$-th word in the story plot of story block \textit{n+1}, 
is used as the target to train the decoder.

The model is built using HuggingFace's GPT-2 implementation
and is trained using AdamW optimizer~\cite{loshchilov2017decoupled} with initial learning rate = $3e-4$, weight decay factor = $1e-4$.
The model is trained for 800,000 steps with batch size = 32.
Top-k sampling is used for generating story plots with top-k = 100, temperature = 0.8, and repetition penalty = 3.0.
Plot lengths are limited to 36 to 76 tokens as the average plot length using GPT-2's tokenizer is 56.

\section{Quality Assessment}
In this study, we examine the quality of plot generation.

\begin{table}[t]
    \centering
    \small
    \begin{tabular}{@{}lcccc@{}}
    \toprule
                        & \textbf{BLEU-4} & \textbf{METEOR} & \textbf{ROUGE-L} & \textbf{ST} \\ \midrule
        \textbf{Rand-History}	& .0012 & .0868 & .1663 & .6101 \\
        \textbf{Rand-Future}	& .0008 & .0881 & .1661 & \textbf{.6106} \\
        \textbf{Fusion-Seq}	    & .0002 & .0741 & .1609 & .5700 \\
        \textbf{P\&W}           & .0005 & .0814 & .1686 & .5836 \\
        \textbf{FGPT-2}	        & \textbf{.0015} & \textbf{.0919} & \textbf{.1744} & .5905 \\ 
    \bottomrule
    \end{tabular}
    \caption{Automatic evaluation metrics for the plot prediction task using Ground-Truth as the reference. Out of the five models, FGPT-2 can generate story plots that are more similar to Ground-Truth.}
    \label{tab:story-plot-auto-eval}
\end{table}

\begin{table}[t]
\small \centering

\begin{tabular}{
    @{}
    l@{}
    S[table-format = 1.3, table-space-text-post = $ $] @{}
    S[table-format = 2.3, table-space-text-post = $^{***}$] @{}
    S[table-format = 2.3, table-space-text-post = $^{***}$] @{}
    S[table-format = 2.3, table-space-text-post = $^{***}$] @{}
    S[table-format = 2.3, table-space-text-post = $^{***}$] @{}
    S[table-format = 2.3, table-space-text-post = $^{***}$] @{}
    @{}
}
\toprule
$\downarrow$ & \multicolumn{1}{c}{\textbf{GT}} & \multicolumn{1}{c}{\textbf{RH}} & \multicolumn{1}{c}{\textbf{RF}} & \multicolumn{1}{c}{\textbf{Fusion}} & \multicolumn{1}{c}{\textbf{P\&W}} & \multicolumn{1}{c@{}}{\textbf{FGPT-2}} \\ \midrule
\textbf{Mean} & 3.091 & 3.586 & 3.528 & 3.733 & 3.741 & 3.321 \\ \midrule

\multicolumn{7}{@{}c@{}}{\textbf{P-values for T-test}} \\ \midrule

                        
\textbf{GT}   & \textendash &  0.001$^{***}$     & 0.001$^{***}$    & 0.001$^{***}$ & 0.001$^{***}$ & .003$^{**}$             \\
\textbf{RH} & \textendash & \textendash   & .437                  & .054               & .039$^{*}$              & 0.001$^{***}$ \\
\textbf{RF}  & \textendash & \textendash   & \textendash  & .006$^{**}$             & .004$^{**}$             & .005$^{*}$              \\
\textbf{Fusion}     & \textendash & \textendash   & \textendash  & \textendash   & .915               & 0.001$^{***}$ \\
\textbf{P\&W}           & \textendash & \textendash   & \textendash  & \textendash   & \textendash   & 0.001$^{***}$ \\

 \bottomrule
\end{tabular}
\caption{Consistency ranking result shows that Ground-Truth $\ll$ FGPT-2 $\ll$ Random-Future $<$ Random-History $\ll$ Fusion-Seq $<$ P\&W, where $\ll$ stands for ``significantly better''. (n=1000, 0.001$^{***}$ is a p-value smaller than 0.001)}
\label{tab:consistency-result}
\end{table}

\begin{table}[t]
\small \centering
\begin{tabular}{
    @{}
    l@{}
    S[table-format = 1.3, table-space-text-post = $ $] @{}
    S[table-format = 2.3, table-space-text-post = $^{***}$] @{}
    S[table-format = 2.3, table-space-text-post = $^{***}$] @{}
    S[table-format = 2.3, table-space-text-post = $^{***}$] @{}
    S[table-format = 2.3, table-space-text-post = $^{***}$] @{}
    S[table-format = 2.3, table-space-text-post = $^{***}$]
    @{}
}
\toprule
$\downarrow$ & \multicolumn{1}{c}{\textbf{GT}} & \multicolumn{1}{c}{\textbf{RH}} & \multicolumn{1}{c}{\textbf{RF}} & \multicolumn{1}{c}{\textbf{Fusion}} & \multicolumn{1}{c}{\textbf{P\&W}} & \multicolumn{1}{c@{}}{\textbf{FGPT-2}} \\ \midrule
\textbf{Mean}      & 3.178                 & 3.402                   & 3.452                  & 3.756               & 3.748               & 3.464               \\ \midrule
\multicolumn{7}{@{}c@{}}{\textbf{P-values for T-test}}                                                                                                                     \\ \midrule

\textbf{GT}   & \textendash & 0.003$^{**}$                & 0.001$^{***}$    & 0.001$^{***}$ & 0.001$^{***}$ & 0.001$^{***}$ \\
\textbf{RH} & \textendash & \textendash   & 0.518                  & 0.001$^{***}$ & 0.001$^{***}$ & 0.414               \\
\textbf{RF}  & \textendash & \textendash   & \textendash  & 0.001$^{***}$ & 0.001$^{***}$ & 0.877               \\
\textbf{Fusion}     & \textendash & \textendash   & \textendash  & \textendash   & 0.915               & 0.001$^{***}$ \\
\textbf{P\&W}           & \textendash & \textendash   & \textendash  & \textendash   & \textendash   & 0.001$^{***}$ \\

\bottomrule
\end{tabular}
\caption{Storiability ranking result shows that Ground-Truth $\ll$ Random-History $<$ Random-Future $<$ FGPT-2 $\ll$ P\&W $<$ Fusion-Seq, where $\ll$ stands for ``significantly better''. (n=1000, 0.001$^{***}$ is a p-value smaller than 0.001)}
\label{tab:storiability-result}
\end{table}

\subsection{Plot Prediction Assessment}
In this plot prediction assessment task, the goal is to 
measure the quality of the automatically generated plots 
by the three models described in the Plot Prediction Section.
We take the human-written plots extracted from the real book as the baselines.
A total of six different models are compared.
\begin{itemize}
    \item \textbf{Ground-Truth (GT).} The gold standard story plot of the story block \textit{n+1}. We expect Ground-Truth to be the upper bound. Note that the story plot is obtained by applying the Matchsum model to extract a three-sentence summary on the target story block. 
    \item \textbf{Random-Future (RF).} The story plot of a randomly selected story block \textit{n+u}, where $5 \leq u \leq 15$. Random-Future is expected to be a strong baseline as it is a follow-up story plot that happened later.
    \item \textbf{Random-History (RH).} The story plot of a randomly selected story block \textit{n-t}, where $10 \leq t \leq 20$. We expect Random-History to be a slightly weaker baseline as it is something that happened.
    \item \textbf{Fusion-Seq.} The fusion-based Seq2seq model described in the Plot Prediction section.
    \item \textbf{P\&W.} The plan-and-write model described in the Plot Prediction section.
    \item \textbf{FGPT-2.} The sequence-to-sequence GPT-2 model guided by frame representations as described in the Plot Prediction section.
\end{itemize}

The evaluation data is built from the \citeauthor{huang-huang-2021-semantic}'s 
Bookcorpus testing set~\cite{huang-huang-2021-semantic} where 958 qualified books are collected.
We randomly sample one story block as the target story block \textit{n} from each book to create a total of 958 testing instances.
In the following sections, we describe how we assess the story plot quality using both
automatic evaluation metrics and human judgments.

\subsection{Automatic Evaluation}
We evaluate the five generated plots by using the Ground-Truth as the reference. 
NLG-eval package~\cite{sharma2017nlgeval} is used and four common metrics,
BLEU-4, METEOR, ROUGE-L, and SkipThought Cosine Similarity (ST), 
are reported in Table~\ref{tab:story-plot-auto-eval}.
The results show that FGPT-2 outperforms other models in the metrics based on token-overlapping, BLEU-4, METEOR, and ROUGE-L.
In the semantic-based metric, ST, the human-written plots, Random-History and Random-Future, still perform better.
However, prior works have shown that automatic evaluation metrics, 
especially token-overlapping metrics, are not entirely suitable for evaluating 
the story generation domain as stories that differed from the target can still be good stories~\cite{hsu2019visual}.
Therefore, we also conduct a human evaluation to measure the quality.

\subsection{Human Evaluation}
We conduct a human evaluation on AMT to evaluate two aspects of the story plot quality, 
consistency and storiability.
In this task, workers are instructed to read a story snippet (20 sentences) along with six follow-up story plots.
We then ask workers to rank the story plots according to consistency and storiability.
Consistency assesses whether the given story plot makes sense in its context (story snippet);
storiability measures whether readers would be curious to read the complete story developed from the given story plot~\cite{roemmele2021inspiration}.
Note that we only ask one single question in a HIT so that workers would not get confused.
To make sure workers spend enough time reading the story snippet and the story plots, we add a 30-second submission lock to the interface.
To alleviate the negative effect and bias caused by the decoding process (such as an unfinished sentence) and control the reading time,
we apply the following three rules to create the evaluation set:
(\emph{i}) all the six story plots have to be within 25-65 words;
(\emph{ii}) the story block \textit{n} has to be within 150-300 words;
(\emph{iii}) the story block \textit{n} and \textit{n+1} do not cross chapters.
Out of the 247 qualified instances, we randomly select 200 instances for evaluation.
For each instance, we collect five assignments which result in a total of 1,000 rankings per aspect.
Given that each HIT contains around 500 words to read, we estimate the task to take around 2 minutes and thus we pay \$0.33 per assignment.

\begin{table*}[t]
    \centering
    \small
    \begin{tabular}{p{17cm}}
    \toprule
        \textbf{Story Snippet.} He looked away, blinking back a tear. The bunker was more centered than Josue had realized, protected on all sides by as much of the manor as possible. He liked that. Anywhere he would walk within the new manor, he would be close to his father's final resting place. John looked down at the square section. "We could rebuild the bunker, if you'd like." Josue was sure, even if they rebuilt it, he would never find the wherewithal to use it. He couldn't think of a more fitting memorial to his father than to leave it the way it was. "Please, don't change a thing." He looked out at the perimeter of the compound. They had almost finished the walls. "Those will protect us." "They will hold for the immediate need." John led Josue through the compound to the eastern hillside overlooking the manor. The shimmer of the obfuscator remained above them. A separate wall surrounded a new plot. He felt a sense of reverence as the Elder led him into the enclosure. In the last rays of the day sun, symmetric monuments reflected pink, planted in neat rows and columns along the hillside. "We took the liberty of adding a memorial garden." John walked to the top of the hill, where a solitary statue stood apart from the others, reminding Josue of a General reviewing his troops. \\ \midrule
        \textbf{Ground-Truth.} Josue wiped his damp cheek. A fire burned in his breast at the very sight of the name. He died doing what he could to protect those he loved." \\ \midrule
        \textbf{Random-History.} Josue tried to remember if he had been on Omri property. Keep away from the mine for a while. His father caught himself and put on a tired smile. \\ \midrule
        \textbf{Random-Future.} Master Hector raised an eyebrow. Josue pushed the thoughts of Timeos out of his mind. Hector pulled a longpole from the rack and threw it to Josue. \\ \midrule
        \textbf{Fusion-Seq.} he shouted. The mob rushed up to the door, and with a sickening clang, the doors began to snap open. With a flash, it was the assassin, the assassin. \\ \midrule
        \textbf{P\&W.} John thought it was a best thing. Bill said, "I have to go home and start a new life." Tom asked, looking at the sign. \\ \midrule
        \textbf{FGPT-2.} John walked back to the edge of the building and looked down. He could see the end of the main staircase on the other side of the wall. It was lined with columns, each leading up to the roof. \\
    \bottomrule
    \end{tabular}
    \caption{Example of the story snippet and the six follow-up story plots.}
    \label{tab:story-plot-example}
\end{table*}

The consistency ranking results in Table~\ref{tab:consistency-result} show that Ground-Truth $\ll$ FGPT-2 $\ll$ Random-Future $<$ Random-History $\ll$ Fusion-Seq $<$ P\&W, where $\ll$ stands for ``significantly better''. 
Surprisingly, FGPT-2 is ranked to be more consistent than the two human-written story plots, Random-Future and Random-History.
The other two story plot generation models, Fusion-Seq and P\&W, are believed to have lower consistency.
Table~\ref{tab:storiability-result} shows the storibility rankings: 
Ground-Truth $\ll$ Random-History $<$ Random-Future $<$ FGPT-2 $\ll$ P\&W $<$ Fusion-Seq. 
Again, Ground-Truth serves as the upper bound and is considered to be the most storiable one.
Although FGPT-2 does not outperform the two human-written story plot baselines here,
it achieves the same level of storiability as them.
We thus conclude that FGPT-2 is a good choice for plot generation as it can produce consistent and storiable story plots.
Table~\ref{tab:story-plot-example} shows an example output of the six models for comparison.


\section{Human Evaluation through a Writing Task}
The quality assessment experiment suggests that FGPT-2 could generate story plots 
that achieve the level of random human-written baselines (Random-Future and Random-History)
in terms of consistency and storability.
To understand how the generated story plots influence people's story writing,
we conduct a preliminary study on AMT using a story continuation task.
In this section, we first describe the study protocol and discuss the result.

\begin{figure}
    \centering
    \includegraphics[width=0.99\linewidth]{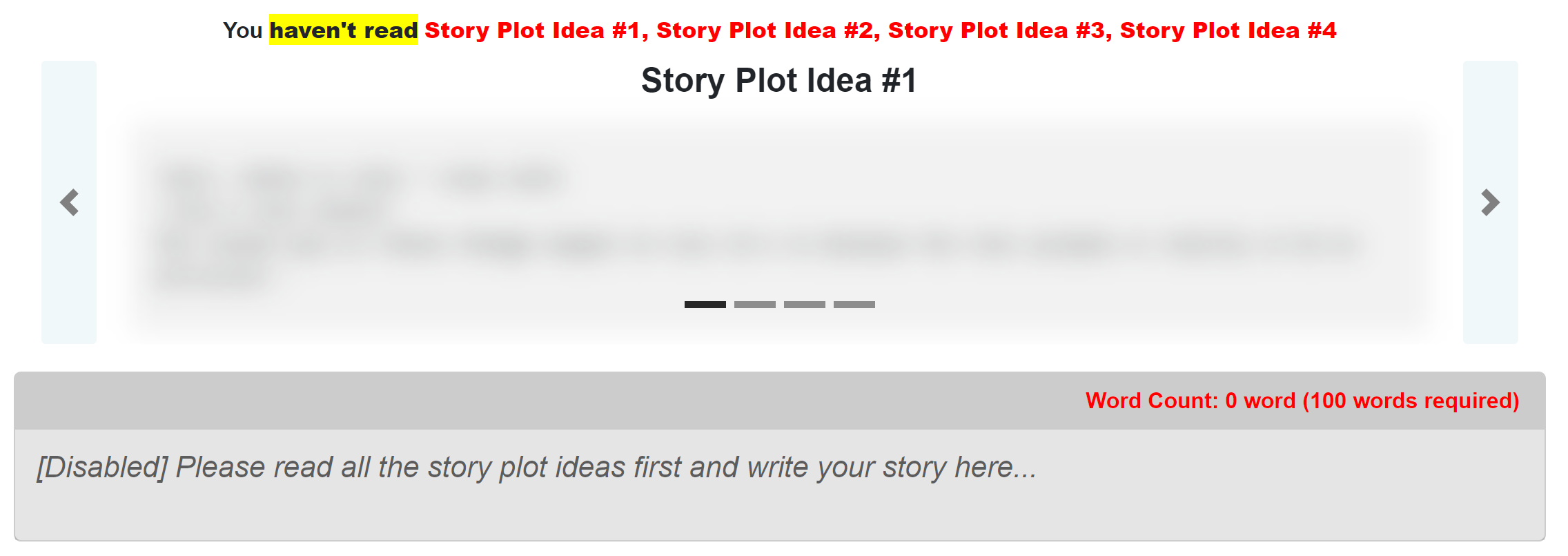}
    \caption{A section of the interface used for the story continuation task. We blur the story plots to prevent workers from copying the exact story plots. Workers would need to click on the story plots to unblur them.}
    \label{fig:writing-study-step-2}
\end{figure}

\subsection{Baselines}
Since writing task is difficult in general, to prevent adding too much workload to workers,
we only compare four different models.
\begin{itemize}
    \item \textbf{Ground-Truth (GT).} As described above.
    \item \textbf{Random-Future (RF).}  As described above.
    \item \textbf{FGPT-2.}  As described above.
    \item \textbf{GPT-3.} We use OpenAI's GPT-3 API~\cite{gpt3-2020} to generate follow-up story plots with the prompt, ``Given the story snippet: [story] Describe a follow-up story arc within 30 words''. We fill in the full story block \textit{n} into [story]. Parameters used was model = text-davinci-002, temperature = 0.95, max-tokens = 76, top-p = 0.95, frequency-penalty = 0.5, presence-penalty = 0.5, and best-of = 5.
\end{itemize}

\subsection{Study Protocol}
In this story continuation task, workers are asked to finish three steps:
(\textit{i}) reading through a story snippet;
(\textit{ii}) writing a 100-word follow-up story to continue the given story snippet with access to story plots; and
(\textit{iii}) answering questions about their experience.

We first show a 20-sentence story snippet (story block \textit{n}) and ask workers to carefully read through it in step (\textit{i}).
In step (\textit{ii}), four story plots generated by four different methods are provided.
Note that the order of the four story plots is randomized.
After finishing reading all the story plots, workers are instructed to write a 100-word story to continue the given story snippet.
As shown in \Cref{fig:writing-study-step-2}, 
to prevent workers from simply copying-and-pasting or typing 
in the exact story plot ideas, we blur the four story plots. 
Workers need to click on the story plots to see the exact texts (the texts would get blurred when the mouse leaves.)
After finishing the writing task, we ask workers to fill-up a short questionnaire regarding their experience with the story plots.
In the questionnaire, we ask seven questions to measure four different aspects:
\begin{enumerate}
    \item \textbf{Inspiringness.} All story plots that inspire the writing (Q1). Multiple selections are allowed.
    \item \textbf{Helpfulness.} The most/least helpful story plot (Q2, Q3).
    \item \textbf{Readability.} The easiest/hardest story plot to comprehend (Q4, Q5).
    \item \textbf{Creativity.} The most/least creative story plot (Q6, Q7).
\end{enumerate}
We add a 180-second submission lock (workers are allowed to submit after 180 seconds) to the interface to ensure workers spend enough time working on the task.
Each task is estimated to be finished within six to nine minutes.
Aiming at paying \$10 dollars per hour, we set the payment for each task to \$1.5 dollars.

As a preliminary study, we only test on five instances. 
We simply take the first five instances from the human evaluation of the quality assessment experiment.
For each instance, we collect five assignments, resulting in a total of 25 stories along with 25 questionnaires.

\begin{table}[t]
\small \centering
\begin{tabular}{
    @{}
    l@{\kern5pt}
    l
    S[table-format = 1.3]
    S[table-format = 1.3]
    S[table-format = 1.3]
    S[table-format = 1.3]
    @{}
}
\toprule
\multicolumn{2}{@{}c}{\textbf{Aspect}}                      & \textbf{GT} & \textbf{RF} & \textbf{FGPT-2} & \textbf{GPT-3}  \\ \midrule
\multicolumn{2}{@{}l}{\textbf{Inspiringness} $\uparrow$}               & 0.294                 & 0.294                  & 0.176         & \textbf{0.647} \\ \midrule
\multirow{3}{*}{\textbf{Helpfulness}} & \textbf{Most} $\uparrow$    & 0.235                 & 0.353                  & 0.059         & 0.353          \\
                                      & \textbf{Least} $\downarrow$ & 0.000                 & 0.294                  & 0.294         & 0.412          \\
                                      & \textbf{Overall} $\uparrow$ & \textbf{0.235}        & 0.059                  & -0.235        & -0.059         \\ \midrule
\multirow{3}{*}{\textbf{Readability}} & \textbf{Easiest} $\uparrow$ & 0.353                 & 0.235                  & 0.176         & 0.235          \\
                                      & \textbf{Hardest} $\downarrow$ & 0.294                 & 0.059                  & 0.471         & 0.176          \\
                                      & \textbf{Overall} $\uparrow$ & 0.059                 & \textbf{0.176}         & -0.294        & 0.059          \\  \midrule
\multirow{3}{*}{\textbf{Creativity}}  & \textbf{Most}    $\uparrow$ & 0.353                 & 0.176                  & 0.000         & 0.471          \\
                                      & \textbf{Least}  $\downarrow$ & 0.176                 & 0.294                  & 0.353         & 0.176          \\
                                      & \textbf{Overall} $\uparrow$ & 0.176                 & -0.118                 & -0.353        & \textbf{0.294} \\ \bottomrule
\end{tabular}
\caption{Questionnaire results from the story continuation task. GPT3 is rated the most inspiring one.}
\label{tab:writing-study}
\end{table}

\subsection{Questionnaire Result}
After obtaining all 25 assignments, we first check out all the written stories.
Despite adding a 180-second submission lock and the blurring function, there are a lot of spamming submissions, 
such as irrelevant texts, random online stories, and random keystrokes.
We read through all the written stories and manually remove the spamming assignments, resulting in a total of 17 assignments remaining.

\Cref{tab:writing-study} shows the questionnaire results.
For inspiringness, as workers are asked to select \textbf{all} story plots that inspired their written story,
we report the percentage of a model being considered inspiring over the 17 assignments.
GPT-3 inspires more than half of the stories (0.647) while FGPT-2 (0.176) is the least inspiring one.

For helpfulness, readability, and creativity, we report the percentage of a model being selected as the most/least helpful, the easiest/hardest readable, and the most/least creative one.
The overall score is computed by Most $-$ Least (or Easiest $-$ Hardest).
Ground-Truth, Random-Future, and GPT-3 are considered the most helpful, the easiest readable, and the most creative, respectively.
We notice that people do not favor FGPT-2 in all three aspects when compared to other baselines.
However, FGPT-2 still inspires 17.6\% of the stories.
Such phenomenon echoes \citeauthor{roemmele2021inspiration}'s finding from the inspiration-through-observation paradigm, 
where human writing would be enhanced by machine-generated texts even though the machine-generated texts are less storible.

We also notice that although GPT-3 gets the highest ``most helpful'' votes, it also gets the highest ``least helpful'' votes.
This would require more analysis to understand why.

\begin{table}[t]
\small \centering
\begin{tabular}{@{}cccccc@{}}
\toprule
                                        & \textbf{GT} & \textbf{RF} & \textbf{FGPT-2} & \textbf{GPT-3}  & \textbf{Random} \\ \midrule
\textbf{Similarity} & 0.816                 & 0.795                  & 0.795         & 0.840     & 0.787   \\ \bottomrule
\end{tabular}
\caption{Semantic similarity between the story plot idea and the written story.}
\label{tab:cosine-similarity}
\end{table}

\begin{table}[]
\centering \small
\begin{tabular}{@{}lrrrr@{}}
\toprule
 & \multicolumn{2}{c}{\textbf{Story Coverage}}                         & \multicolumn{2}{c}{\textbf{Plot Coverage}}                          \\ \cmidrule{2-5}
 & \multicolumn{1}{c}{\textbf{Mean}} & \multicolumn{1}{c}{\textbf{CI}} & \multicolumn{1}{c}{\textbf{Mean}} & \multicolumn{1}{c}{\textbf{CI}} \\ \midrule
\textbf{GT}     & 0.198 & {[}0.163, 0.233{]} & 0.530 & {[}0.473, 0.587{]} \\
\textbf{RF}     & 0.193 & {[}0.164, 0.222{]} & 0.536 & {[}0.475, 0.598{]} \\
\textbf{FGPT-2} & 0.163 & {[}0.145, 0.182{]} & 0.484 & {[}0.429, 0.539{]} \\
\textbf{GPT-3}  & 0.170 & {[}0.149, 0.190{]} & 0.498 & {[}0.441, 0.555{]} \\ 
\textbf{Random} & 0.151 & {[}0.149, 0.153{]} & 0.450 & {[}0.445, 0.455{]} \\ \bottomrule
\end{tabular}
\caption{GT and RF contribute more in token level.}
\label{tab:alignment}
\end{table}

\subsection{How do story plots affect story writing?}
The questionnaire serves as self-reported results.
To understand how the story plots influence story writing, 
we analyze the relationship between the story plots and the follow-up story written by workers.
Note that each HIT assignment comes with one human-written follow-up story and four machine-generated story plots.
Using this data, we measure two aspects: (\textit{i}) semantic similarity and (\textit{ii}) token overlap.

\paragraph{Semantic Similarity.}
We encode the follow-up stories and the story plots using Sentence-BERT~\cite{reimers-gurevych-2019-sentence}\footnote{\texttt{sentence-transformers/sentence-t5-base}}.
Cosine similarity is then used to compute the semantic similarity between a follow-up story and a story plot.
To get a better sense of this cosine similarity value, we also include a Random baseline for comparison.
The Random baseline reports the semantic similarity between the follow-up stories and a set of randomly sampled 40-word paragraphs.
We choose 40 words to match the length of the story plots.
These paragraphs are randomly selected from NLTK's Gutenberg corpus.
We start with 3,400 random paragraphs (two sentences) and remove those shorter than 30 words or longer than 50 words.
A total of 1,113 valid short paragraphs are then included in this random set.

As shown in \Cref{tab:cosine-similarity}, story plots generated by GPT-3 have the highest semantic similarity (0.840)
with the human-written follow-up stories, suggesting that workers do get inspiration from GPT-3.
Ground-Truth (0.816), Random-Future (0.795), and FGPT-2 (0.795) are slightly lower.
However, all of the models are higher than the Random baseline suggesting that workers do get inspired more or less, 
which again echos the inspiration-through-observation paradigm~\cite{roemmele2021inspiration}.

\paragraph{Token Overlap.}
To understand the effect in the token level, we use awesome-align~\cite{dou2021word} 
to get the token alignment between the follow-up story and the story plots.
Awesome-align computes the similarity over tokens' contextual embeddings to assign alignments.
Compared to the exact token overlap, awesome-align provides a soft overlap where semantically similar words can be identified.
Upon getting the alignment, we compute two scores.
\begin{enumerate}
    \item \textbf{Story Coverage}: the percentage of the follow-up stories that can be found in the story plots. This also means the amount of information contributed by the story plots.
    \item \textbf{Plot Coverage}: the percentage of the story plots that can be found in the follow-up stories. This also indicates the helpfulness of the story plots.
\end{enumerate}
Note that when computing the coverage score, we exclude punctuation and stop words.
Again, we add a Random baseline to help us interpret the scores.
As shown in~\Cref{tab:alignment}, Ground-Truth and Random-Future help more at the token level.
This is probably because they provide wording and terms more useful in the context.
Although FGPT-2 is less helpful compared to other strong baselines, it still somewhat affects workers' writing.

The difference between semantic helpfulness and token-level helpfulness probably causes workers to vote GPT-3 as the most helpful but also the least helpful.

\section{Discussion}

The preliminary experiment on the story continuation task somewhat suggests that the proposed FGPT-2 approach only provides limited help in real writing tasks. We identify a few possible reasons.

\paragraph{Limitations of our story plot formulation.}
To provide support in long stories, we extract only three sentences from a story block to form a story plot.
We expect story plots to capture all the essential information and represent the story block, but much of the information is indeed missing.
This is probably caused by the use of the extractive summarization method, which essentially selects a few sentences from the story block. 
Therefore, if we would like to include all the essential information, the story would need to contain a few very informative sentences.
This is not the case for stories.
In the future, we would explore other ways to condense information for a story block.

\paragraph{The study on the short story writing task might not capture the difficulty of novel writing.}
The proposed method is built for stories, such as novels, that are long enough to break recent story generation models. 
However, conducting experiments with people on such long texts is hard.
To evaluate the system in the desired context, we will conduct a formal user study in the future.

\section{Conclusion}
In this paper, we generate short future story plots as follow-up story ideas to help writers continue their story.
Results on AMT confirm that using FGPT-2 to serve as the plot prediction model yields plots that are
significantly more consistent than the ones generated by the two human-written baselines, Random-History and Random-Future.
When comparing the storiability, how appealing a story plot is for readers, FGPT-2 is competitive to the two random baselines.
A preliminary human study with a story continuation task suggests that FGPT-2 could positively affect story writing but there are still difficulties to overcome.
In the future, we will (\textit{i}) explore better ways to build story plots and 
(\textit{ii}) integrate the proposed function to a real editor (\eg, Google Docs) and conduct studies to measure whether writers can get benefits for fiction writing in practice.

\section{Acknowledgments}
The authors extend their heartfelt appreciation to the late Dr. Arzoo Katiyar for her unwavering support and insightful suggestions throughout the course of this project.
The authors also acknowledge the valuable contributions of the anonymous reviewers, who provided constructive feedback to enhance the quality of this work.

\bibliography{AnonymousSubmission/bib/aaai23}

\begin{thebibliography}{47}
\providecommand{\natexlab}[1]{#1}

\bibitem[{Akoury et~al.(2020)Akoury, Wang, Whiting, Hood, Peng, and
  Iyyer}]{akoury-etal-2020-storium}
Akoury, N.; Wang, S.; Whiting, J.; Hood, S.; Peng, N.; and Iyyer, M. 2020.
\newblock {STORIUM}: {A} {D}ataset and {E}valuation {P}latform for
  {M}achine-in-the-{L}oop {S}tory {G}eneration.
\newblock In \emph{Proceedings of the 2020 Conference on Empirical Methods in
  Natural Language Processing (EMNLP)}, 6470--6484. Online: Association for
  Computational Linguistics.

\bibitem[{Bamman, O{'}Connor, and Smith(2013)}]{bamman-etal-2013-learning}
Bamman, D.; O{'}Connor, B.; and Smith, N.~A. 2013.
\newblock Learning Latent Personas of Film Characters.
\newblock In \emph{Proceedings of the 51st Annual Meeting of the Association
  for Computational Linguistics (Volume 1: Long Papers)}, 352--361. Sofia,
  Bulgaria: Association for Computational Linguistics.

\bibitem[{Bird, Klein, and Loper(2009)}]{BirdKleinLoper09}
Bird, S.; Klein, E.; and Loper, E. 2009.
\newblock \emph{{Natural Language Processing with Python}}.
\newblock O'Reilly Media.

\bibitem[{Brown et~al.(2020)Brown, Mann, Ryder, Subbiah, Kaplan, Dhariwal,
  Neelakantan, Shyam, Sastry, Askell, Agarwal, Herbert-Voss, Krueger, Henighan,
  Child, Ramesh, Ziegler, Wu, Winter, Hesse, Chen, Sigler, Litwin, Gray, Chess,
  Clark, Berner, McCandlish, Radford, Sutskever, and Amodei}]{gpt3-2020}
Brown, T.; Mann, B.; Ryder, N.; Subbiah, M.; Kaplan, J.~D.; Dhariwal, P.;
  Neelakantan, A.; Shyam, P.; Sastry, G.; Askell, A.; Agarwal, S.;
  Herbert-Voss, A.; Krueger, G.; Henighan, T.; Child, R.; Ramesh, A.; Ziegler,
  D.; Wu, J.; Winter, C.; Hesse, C.; Chen, M.; Sigler, E.; Litwin, M.; Gray,
  S.; Chess, B.; Clark, J.; Berner, C.; McCandlish, S.; Radford, A.; Sutskever,
  I.; and Amodei, D. 2020.
\newblock Language Models are Few-Shot Learners.
\newblock In Larochelle, H.; Ranzato, M.; Hadsell, R.; Balcan, M.; and Lin, H.,
  eds., \emph{Advances in Neural Information Processing Systems}, volume~33,
  1877--1901. Curran Associates, Inc.

\bibitem[{Chakrabarty, Padmakumar, and He(2022)}]{chakrabarty2022help}
Chakrabarty, T.; Padmakumar, V.; and He, H. 2022.
\newblock Help me write a poem: Instruction Tuning as a Vehicle for
  Collaborative Poetry Writing.
\newblock In \emph{Proceedings of the 2022 Conference on Empirical Methods in
  Natural Language Processing}.

\bibitem[{Chung et~al.(2022)Chung, Kim, Yoo, Lee, Adar, and
  Chang}]{chung2022talebrush}
Chung, J. J.~Y.; Kim, W.; Yoo, K.~M.; Lee, H.; Adar, E.; and Chang, M. 2022.
\newblock TaleBrush: sketching stories with generative pretrained language
  models.
\newblock In \emph{Proceedings of the 2022 CHI Conference on Human Factors in
  Computing Systems}, 1--19.

\bibitem[{Clark et~al.(2018)Clark, Ross, Tan, Ji, and
  Smith}]{Clark:2018:CWM:3172944.3172983}
Clark, E.; Ross, A.~S.; Tan, C.; Ji, Y.; and Smith, N.~A. 2018.
\newblock Creative Writing with a Machine in the Loop: Case Studies on Slogans
  and Stories.
\newblock In \emph{23rd International Conference on Intelligent User
  Interfaces}, IUI '18, 329--340. New York, NY, USA: ACM.
\newblock ISBN 978-1-4503-4945-1.

\bibitem[{Dou and Neubig(2021)}]{dou2021word}
Dou, Z.-Y.; and Neubig, G. 2021.
\newblock Word Alignment by Fine-tuning Embeddings on Parallel Corpora.
\newblock In \emph{Conference of the European Chapter of the Association for
  Computational Linguistics (EACL)}.

\bibitem[{Fan, Lewis, and Dauphin(2018)}]{fan-etal-2018-hierarchical}
Fan, A.; Lewis, M.; and Dauphin, Y. 2018.
\newblock Hierarchical Neural Story Generation.
\newblock In \emph{Proceedings of the 56th Annual Meeting of the Association
  for Computational Linguistics (Volume 1: Long Papers)}, 889--898. Melbourne,
  Australia: Association for Computational Linguistics.

\bibitem[{Fan, Lewis, and Dauphin(2019)}]{fan-etal-2019-strategies}
Fan, A.; Lewis, M.; and Dauphin, Y. 2019.
\newblock Strategies for Structuring Story Generation.
\newblock In \emph{Proceedings of the 57th Annual Meeting of the Association
  for Computational Linguistics}, 2650--2660. Florence, Italy: Association for
  Computational Linguistics.

\bibitem[{Flower and Hayes(1981)}]{flower1981cognitive}
Flower, L.; and Hayes, J.~R. 1981.
\newblock A cognitive process theory of writing.
\newblock \emph{College composition and communication}, 32(4): 365--387.

\bibitem[{Gabriel, Chen, and Nichols(2015)}]{gabriel2015inkwell}
Gabriel, R.~P.; Chen, J.; and Nichols, J. 2015.
\newblock InkWell: A Creative Writer's Creative Assistant.
\newblock In \emph{Proceedings of the 2015 ACM SIGCHI Conference on Creativity
  and Cognition}, 93--102. ACM.

\bibitem[{Gero and Chilton(2019)}]{10.1145/3290605.3300526}
Gero, K.~I.; and Chilton, L.~B. 2019.
\newblock Metaphoria: An Algorithmic Companion for Metaphor Creation.
\newblock In \emph{Proceedings of the 2019 CHI Conference on Human Factors in
  Computing Systems}, CHI '19, 1–12. New York, NY, USA: Association for
  Computing Machinery.
\newblock ISBN 9781450359702.

\bibitem[{Gero, Liu, and Chilton(2022)}]{10.1145/3532106.3533533}
Gero, K.~I.; Liu, V.; and Chilton, L. 2022.
\newblock Sparks: Inspiration for Science Writing Using Language Models.
\newblock In \emph{Designing Interactive Systems Conference}, DIS '22,
  1002–1019. New York, NY, USA: Association for Computing Machinery.
\newblock ISBN 9781450393584.

\bibitem[{Goldfarb-Tarrant et~al.(2020)Goldfarb-Tarrant, Chakrabarty,
  Weischedel, and Peng}]{goldfarb-tarrant-etal-2020-content}
Goldfarb-Tarrant, S.; Chakrabarty, T.; Weischedel, R.; and Peng, N. 2020.
\newblock Content Planning for Neural Story Generation with Aristotelian
  Rescoring.
\newblock In \emph{Proceedings of the 2020 Conference on Empirical Methods in
  Natural Language Processing (EMNLP)}, 4319--4338. Online: Association for
  Computational Linguistics.

\bibitem[{Hsu et~al.(2019)Hsu, Huang, Hsu, and Huang}]{hsu2019visual}
Hsu, T.-Y.; Huang, C.-Y.; Hsu, Y.-C.; and Huang, T.-H. 2019.
\newblock Visual Story Post-Editing.
\newblock In \emph{Proceedings of the 57th Annual Meeting of the Association
  for Computational Linguistics}, 6581--6586.

\bibitem[{Hua and Wang(2020)}]{hua2020pair}
Hua, X.; and Wang, L. 2020.
\newblock {PAIR}: Planning and Iterative Refinement in Pre-trained Transformers
  for Long Text Generation.
\newblock In \emph{Proceedings of the 2020 Conference on Empirical Methods in
  Natural Language Processing (EMNLP)}, 781--793. Online: Association for
  Computational Linguistics.

\bibitem[{Huang, Huang, and Huang(2020)}]{huang2020heteroglossia}
Huang, C.-Y.; Huang, S.-H.; and Huang, T.-H.~K. 2020.
\newblock Heteroglossia: In-Situ Story Ideation with the Crowd.
\newblock In \emph{Proceedings of the 2020 CHI Conference on Human Factors in
  Computing Systems}, 1--12.

\bibitem[{Huang and Huang(2021)}]{huang-huang-2021-semantic}
Huang, C.-Y.; and Huang, T.-H. 2021.
\newblock Semantic Frame Forecast.
\newblock In \emph{Proceedings of the 2021 Conference of the North American
  Chapter of the Association for Computational Linguistics: Human Language
  Technologies}, 2702--2713. Online: Association for Computational Linguistics.

\bibitem[{Ippolito et~al.(2022)Ippolito, Yuan, Coenen, and
  Burnam}]{ippolito2022creative}
Ippolito, D.; Yuan, A.; Coenen, A.; and Burnam, S. 2022.
\newblock Creative Writing with an AI-Powered Writing Assistant: Perspectives
  from Professional Writers.
\newblock \emph{arXiv preprint arXiv:2211.05030}.

\bibitem[{Keskar et~al.(2019)Keskar, McCann, Varshney, Xiong, and
  Socher}]{keskar2019ctrl}
Keskar, N.~S.; McCann, B.; Varshney, L.~R.; Xiong, C.; and Socher, R. 2019.
\newblock Ctrl: A conditional transformer language model for controllable
  generation.
\newblock \emph{arXiv preprint arXiv:1909.05858}.

\bibitem[{Kingma and Ba(2015)}]{kingma2014adam}
Kingma, D.~P.; and Ba, J. 2015.
\newblock Adam: {A} Method for Stochastic Optimization.
\newblock In Bengio, Y.; and LeCun, Y., eds., \emph{3rd International
  Conference on Learning Representations, {ICLR} 2015, San Diego, CA, USA, May
  7-9, 2015, Conference Track Proceedings}.

\bibitem[{Kry{\'s}ci{\'n}ski et~al.(2021)Kry{\'s}ci{\'n}ski, Rajani, Agarwal,
  Xiong, and Radev}]{kryscinski2021booksum}
Kry{\'s}ci{\'n}ski, W.; Rajani, N.; Agarwal, D.; Xiong, C.; and Radev, D. 2021.
\newblock BookSum: A Collection of Datasets for Long-form Narrative
  Summarization.
\newblock \emph{arXiv preprint arXiv:2105.08209}.

\bibitem[{Lee, Liang, and Yang(2022)}]{10.1145/3491102.3502030}
Lee, M.; Liang, P.; and Yang, Q. 2022.
\newblock CoAuthor: Designing a Human-AI Collaborative Writing Dataset for
  Exploring Language Model Capabilities.
\newblock In \emph{Proceedings of the 2022 CHI Conference on Human Factors in
  Computing Systems}, CHI '22. New York, NY, USA: Association for Computing
  Machinery.
\newblock ISBN 9781450391573.

\bibitem[{Li et~al.(2013)Li, Lee-Urban, Johnston, and Riedl}]{li2013story}
Li, B.; Lee-Urban, S.; Johnston, G.; and Riedl, M. 2013.
\newblock Story generation with crowdsourced plot graphs.
\newblock In \emph{Twenty-Seventh AAAI Conference on Artificial Intelligence}.

\bibitem[{Li and Riedl(2015)}]{li2015scheherazade}
Li, B.; and Riedl, M. 2015.
\newblock Scheherazade: Crowd-powered interactive narrative generation.
\newblock In \emph{Twenty-Ninth AAAI Conference on Artificial Intelligence}.

\bibitem[{Lita et~al.(2003)Lita, Ittycheriah, Roukos, and
  Kambhatla}]{lita2003truecasing}
Lita, L.~V.; Ittycheriah, A.; Roukos, S.; and Kambhatla, N. 2003.
\newblock Truecasing.
\newblock In \emph{Proceedings of the 41st Annual Meeting of the Association
  for Computational Linguistics}, 152--159.

\bibitem[{Loshchilov and Hutter(2019)}]{loshchilov2017decoupled}
Loshchilov, I.; and Hutter, F. 2019.
\newblock Decoupled Weight Decay Regularization.
\newblock In \emph{International Conference on Learning Representations}.

\bibitem[{Martin et~al.(2018)Martin, Ammanabrolu, Wang, Hancock, Singh,
  Harrison, and Riedl}]{martin2018event}
Martin, L.~J.; Ammanabrolu, P.; Wang, X.; Hancock, W.; Singh, S.; Harrison, B.;
  and Riedl, M.~O. 2018.
\newblock Event representations for automated story generation with deep neural
  nets.
\newblock In \emph{Thirty-Second AAAI Conference on Artificial Intelligence}.

\bibitem[{Mirowski et~al.(2022)Mirowski, Mathewson, Pittman, and
  Evans}]{mirowski2022co}
Mirowski, P.; Mathewson, K.~W.; Pittman, J.; and Evans, R. 2022.
\newblock Co-writing screenplays and theatre scripts with language models: An
  evaluation by industry professionals.
\newblock \emph{arXiv preprint arXiv:2209.14958}.

\bibitem[{Mori et~al.(2022)Mori, Yamane, Shimizu, Mukuta, and
  Harada}]{DBLP:journals/corr/abs-2202-13151}
Mori, Y.; Yamane, H.; Shimizu, R.; Mukuta, Y.; and Harada, T. 2022.
\newblock COMPASS: a Creative Support System that Alerts Novelists to the
  Unnoticed Missing Contents.
\newblock \emph{CoRR}, abs/2202.13151.

\bibitem[{Mostafazadeh et~al.(2016)Mostafazadeh, Chambers, He, Parikh, Batra,
  Vanderwende, Kohli, and Allen}]{mostafazadeh-etal-2016-corpus}
Mostafazadeh, N.; Chambers, N.; He, X.; Parikh, D.; Batra, D.; Vanderwende, L.;
  Kohli, P.; and Allen, J. 2016.
\newblock A Corpus and Cloze Evaluation for Deeper Understanding of Commonsense
  Stories.
\newblock In \emph{NAACL'16}, 839--849. San Diego, California: ACL.

\bibitem[{Mostafazadeh et~al.(2020)Mostafazadeh, Kalyanpur, Moon, Buchanan,
  Berkowitz, Biran, and Chu-Carroll}]{mostafazadeh-etal-2020-glucose}
Mostafazadeh, N.; Kalyanpur, A.; Moon, L.; Buchanan, D.; Berkowitz, L.; Biran,
  O.; and Chu-Carroll, J. 2020.
\newblock {GLUCOSE}: {G}enera{L}ized and {CO}ntextualized Story Explanations.
\newblock In \emph{Proceedings of the 2020 Conference on Empirical Methods in
  Natural Language Processing (EMNLP)}, 4569--4586. Online: Association for
  Computational Linguistics.

\bibitem[{Padmakumar and He(2022)}]{padmakumar-he-2022-machine}
Padmakumar, V.; and He, H. 2022.
\newblock Machine-in-the-Loop Rewriting for Creative Image Captioning.
\newblock In \emph{Proceedings of the 2022 Conference of the North American
  Chapter of the Association for Computational Linguistics: Human Language
  Technologies}, 573--586. Seattle, United States: Association for
  Computational Linguistics.

\bibitem[{Peng et~al.(2018)Peng, Ghazvininejad, May, and
  Knight}]{peng-etal-2018-towards}
Peng, N.; Ghazvininejad, M.; May, J.; and Knight, K. 2018.
\newblock Towards Controllable Story Generation.
\newblock In \emph{Proceedings of the First Workshop on Storytelling}, 43--49.
  New Orleans, Louisiana: Association for Computational Linguistics.

\bibitem[{Radford et~al.(2019)Radford, Wu, Child, Luan, Amodei, and
  Sutskever}]{radford2019language}
Radford, A.; Wu, J.; Child, R.; Luan, D.; Amodei, D.; and Sutskever, I. 2019.
\newblock Language models are unsupervised multitask learners.
\newblock \emph{OpenAI Blog}, 1(8).

\bibitem[{Reimers and Gurevych(2019)}]{reimers-gurevych-2019-sentence}
Reimers, N.; and Gurevych, I. 2019.
\newblock Sentence-{BERT}: Sentence Embeddings using {S}iamese {BERT}-Networks.
\newblock In \emph{Proceedings of the 2019 Conference on Empirical Methods in
  Natural Language Processing and the 9th International Joint Conference on
  Natural Language Processing (EMNLP-IJCNLP)}, 3982--3992. Hong Kong, China:
  Association for Computational Linguistics.

\bibitem[{Riedl and Young(2010)}]{riedl2010narrative}
Riedl, M.~O.; and Young, R.~M. 2010.
\newblock Narrative planning: Balancing plot and character.
\newblock \emph{Journal of Artificial Intelligence Research}, 39: 217--268.

\bibitem[{Roemmele(2021)}]{roemmele2021inspiration}
Roemmele, M. 2021.
\newblock Inspiration through Observation: Demonstrating the Influence of
  Automatically Generated Text on Creative Writing.
\newblock In \emph{12th International Conference on Computational Creativity}.

\bibitem[{Roemmele and Gordon(2015)}]{roemmele2015creative}
Roemmele, M.; and Gordon, A.~S. 2015.
\newblock Creative help: a story writing assistant.
\newblock In \emph{International Conference on Interactive Digital
  Storytelling}, 81--92. Springer.

\bibitem[{Rose et~al.(2010)Rose, Engel, Cramer, and Cowley}]{RAKE2010}
Rose, S.; Engel, D.; Cramer, N.; and Cowley, W. 2010.
\newblock \emph{Automatic Keyword Extraction from Individual Documents}, 1 --
  20.
\newblock ISBN 9780470689646.

\bibitem[{See et~al.(2019)See, Pappu, Saxena, Yerukola, and
  Manning}]{see-etal-2019-massively}
See, A.; Pappu, A.; Saxena, R.; Yerukola, A.; and Manning, C.~D. 2019.
\newblock Do Massively Pretrained Language Models Make Better Storytellers?
\newblock In \emph{Proceedings of the 23rd Conference on Computational Natural
  Language Learning (CoNLL)}, 843--861. Hong Kong, China: Association for
  Computational Linguistics.

\bibitem[{Sharma et~al.(2017)Sharma, El~Asri, Schulz, and
  Zumer}]{sharma2017nlgeval}
Sharma, S.; El~Asri, L.; Schulz, H.; and Zumer, J. 2017.
\newblock Relevance of Unsupervised Metrics in Task-Oriented Dialogue for
  Evaluating Natural Language Generation.
\newblock \emph{CoRR}, abs/1706.09799.

\bibitem[{Swanson et~al.(2021)Swanson, Mathewson, Pietrzak, Chen, and
  Dinalescu}]{swanson-etal-2021-story}
Swanson, B.; Mathewson, K.; Pietrzak, B.; Chen, S.; and Dinalescu, M. 2021.
\newblock Story Centaur: Large Language Model Few Shot Learning as a Creative
  Writing Tool.
\newblock In \emph{Proceedings of the 16th Conference of the European Chapter
  of the Association for Computational Linguistics: System Demonstrations},
  244--256. Online: Association for Computational Linguistics.

\bibitem[{Xu et~al.(2018)Xu, Ren, Zhang, Zeng, Cai, and Sun}]{xu2018skeleton}
Xu, J.; Ren, X.; Zhang, Y.; Zeng, Q.; Cai, X.; and Sun, X. 2018.
\newblock A Skeleton-Based Model for Promoting Coherence Among Sentences in
  Narrative Story Generation.
\newblock In \emph{Proceedings of the 2018 Conference on Empirical Methods in
  Natural Language Processing}, 4306--4315. Brussels, Belgium: Association for
  Computational Linguistics.

\bibitem[{Yao et~al.(2019)Yao, Peng, Weischedel, Knight, Zhao, and
  Yan}]{yao2019plan}
Yao, L.; Peng, N.; Weischedel, R.; Knight, K.; Zhao, D.; and Yan, R. 2019.
\newblock Plan-and-write: Towards better automatic storytelling.
\newblock In \emph{Proceedings of the AAAI Conference on Artificial
  Intelligence}, volume~33, 7378--7385.

\bibitem[{Zhong et~al.(2020)Zhong, Liu, Chen, Wang, Qiu, and
  Huang}]{zhong2020extractive}
Zhong, M.; Liu, P.; Chen, Y.; Wang, D.; Qiu, X.; and Huang, X. 2020.
\newblock Extractive Summarization as Text Matching.
\newblock In \emph{Proceedings of the 58th Annual Meeting of the Association
  for Computational Linguistics}, 6197--6208. Online: Association for
  Computational Linguistics.

\end{thebibliography}

\end{document}